\documentclass{article}

   \usepackage[preprint]{neurips_2024}

\usepackage[utf8]{inputenc} 
\usepackage[T1]{fontenc}    
\usepackage{hyperref}       
\usepackage{url}            
\usepackage{natbib}
\usepackage{booktabs}       
\usepackage{amsfonts}       
\usepackage{nicefrac}       
\usepackage{microtype}      
\usepackage{xcolor}         
\usepackage{amsmath}
\usepackage{graphicx, color}
\graphicspath{{images/}}
\usepackage{multirow}

\title{Human Motion Synthesis: A Diffusion Approach for Motion Stitching and In-Betweening}

\author{%
  Michael~Adewole\thanks{corresponding authors - michaseyi@gmail.com and giwaoluwaseyi475@gmail.com.} \\
  Department of Computer Engineering\\
  Olabisi Onabanjo University\\
  Ago-Iwoye, Ogun State\\
  \texttt{michaseyi@gmail.com} \\
   \AND
   Oluwaseyi~Giwa \\
   African Institute for Mathematical Sciences\\
   South Africa\\
   \texttt{oluwaseyi@aims.ac.za} \\
   \AND
   Favour~Nerrise \\
   Department of Electrical Engineering \\
   Stanford University \\
   \texttt{fnerrise@stanford.edu} \\
   \AND
   Martins~Osifeko \\
   Department of Computer Engineering \\
   Olabisi Onabanjo University \\
   \texttt{osifeko.martins@oouagoiwoye.edu.ng} \\
   \AND
   Ajibola~Oyedeji \\
   Department of Computer Engineering \\
   Olabisi Onabanjo University \\
   \texttt{oyedeji.ajibola@oouagoiwoye.edu.ng} \\
}

\begin{document}

\maketitle

\begin{abstract}
  Human motion generation is an important area of research in many fields. In this work, we tackle the problem of motion stitching and in-betweening. Current methods either require manual efforts, or are incapable of handling longer sequences. To address these challenges, we propose a diffusion model with a transformer-based denoiser to generate realistic human motion. Our method demonstrated strong performance in generating in-betweening sequences, transforming a variable number of input poses into smooth and realistic motion sequences consisting of 75 frames at 15 fps, resulting in a total duration of 5 seconds. We present the performance evaluation of our method using quantitative metrics such as Frechet Inception Distance (FID), Diversity, and Multimodality, along with visual assessments of the generated outputs. 
\end{abstract}

\section{Introduction}

Human motion data are important, perhaps evident by its extensive usage across various fields. Human motion generation techniques play important roles from driving character animation movies and video games to robotics striving for a more natural and human-like feel. Motion capture devices and hand-sculpted human motion sequences are common ways to acquire human motion. However, due to the high cost of quality motion capture devices, the skills required to keyframe human motion manually, and their flexibility and limitations, it is not feasible for many applications. The demand for high-quality human motion data has precipitated extensive research on human motion generation, with the primary goal of generating realistic and natural human motion.\newline
Over the years, there have been numerous work leveraging the generative capabilities of artificial intelligence for this task. In particular, we have seen the use of neural network-based models such as diffusion transformer models \citep{Shi2023}, generative adversarial networks (GAN) \citep{Lin2018}, variable autoencoders (VAE) \citep{Habibie2017}, convolutional neural networks (CNN) \citep{Zhou2019, Pavllo2019} and recurrent neural networks (RNN) \citep{Martinez2017, Zhou2018, Pavllo2018} producing promising results. These approaches offer the potential to overcome the limitations of traditional methods, making high-quality human motion more accessible for many applications.

Despite recent advances in human motion generation, there is relatively little research focused on motion stitching. Motion stitching involves generating a realistic motion sequence that passes through given keyframes. These keyframes can appear at any point in the sequence. Although existing studies address continuous motion generation from prior motion data, few explicitly tackle this challenge. For example, \citep{Martinez2017} proposed a sequence-to-sequence RNN architecture with residual connections to enhance short-term human motion prediction by directly modeling motion velocities. The sampling-based loss function was used to ensure the model recovered from its prediction errors during training. QuaterNet \citep{Pavllo2019}, an extension of earlier \citep{Pavllo2018}, uses a CNN over its original RNN approach and focuses on quaternions for rotation representation, eliminating discontinuities and singularities that can hinder training. QuaterNet uses a positional geometric loss function, the predicted rotations are converted to joint positions using forward kinematics. However, these models are limited as they require all motion frames to follow each other and be positioned at the start of the sequence.

\citep{Harvey2021} addressed this challenge by proposing the use of an adversarial RNN architecture with additive embedding modifiers to handle varying transition lengths. It also used scheduled target noise vectors for a diverse generation of realistic motions. However, their method operated primarily in the latent space of the RNN which could limit its ability to fully leverage the explicit temporal relationships that exist between motion frames.

To address these challenges, we propose a diffusion model approach. First, we pass the input motion frames encoded with their position in the motion sequence and the current diffusion step into an encoder transformer. Here, we capture the relationships between motion frames with the help of the self-attention mechanism as described in \citep{Vaswani2017}. Next, the encoder output is used together with an initial Gaussian noise as input  to another encoder transformer to predict the clean motion at the given time step. The role of the transformer is to serve as a denoiser as described in \citep{Ho2020}. We incorporate the idea of predicting clean data instead of noise from \citep{Shi2023}. The output of this transformer is then fed to a noise scheduler which adds random Gaussian noise based on the current time step to retrieve the noisy input for that time step. This noisy input is then returned in place of the initial Gaussian noise. The entire process is repeated for a predefined number of times.

Our contributions are as follows:
\begin{itemize}
    \item We developed a diffusion model that generates realistic human motion, filling missing motion frames within a given sequence.
    \item We demonstrated the effectiveness of our method through extensive experiments on short-term and long-term motion generation tasks.
\end{itemize}

\begin{figure}[htbp]
    \centering
    \includegraphics[width=0.5\textwidth]{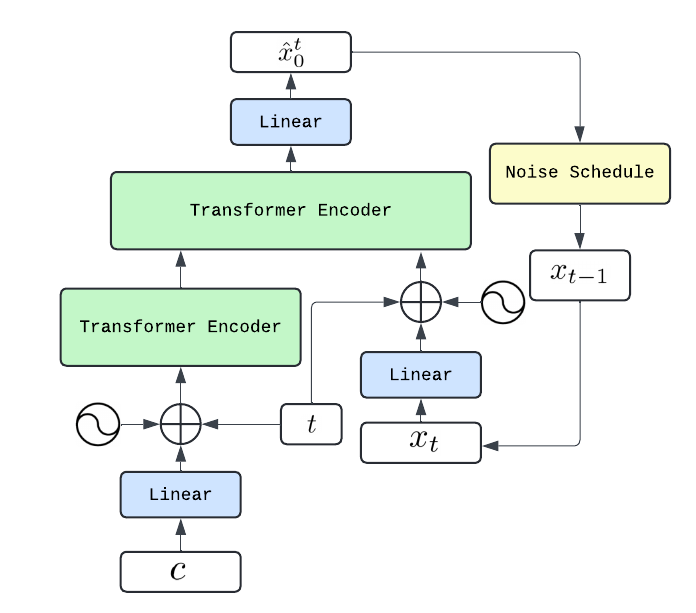}
    \caption{The workflow of our approach. Contextual information is extracted from the input poses \(c\) using a transformer encoder. The output is used to transform noisy motion data \(x_t\) to clean motion \(\hat{x}_0^t\) using another transformer encoder. The clean motion goes through the noise schedule to generate the noisy data \(x_{t-1}\) for the next timestep. This process is repeated for a predefined number of iterations.}
    \label{fig:architecture}
\end{figure}

\section{Related Works}
\subsection{Human Motion Generation}
Prior work on human motion generation can be classified according to the input condition used for motion generation. These classifications include text-to-motion, action class-to-motion, prior motion-to-motion, and video-to-motion.

Transformation of text into motion involves generating motion from a text description. An encoder like CLIP \citep{Radford2021} is often used to encode the input. \citep{Chen2022} used VAEs to learn a representation of human motion, then performed diffusion on this latent space representation instead of the raw motion data. MotionCLIP \citep{Tevet2022} also used autoencoders, but instead mapped the CLIP representation of text to motion. MotionGPT \citep{Zhang2024} employed LLMs fine-tuned with LoRA \citep{Hu2021} for motion generation. Motion generation tasks are fed into it as a prompt. \citep{Guo2022} used autoencoders to generate motion sequences with dynamic lengths from text by also predicting the length from the text. They also proposed the HumanML3D dataset. FG-MDM \citep{Shi2023} further improved the generalization capabilities of these models by conditioning the input with LLMs. FG-MDM used LLMs to expand the text description of motion from the HumanML3D dataset \citep{Guo2022} by including additional descriptions of several body parts through the motion encoded with CLIP.

Action class-to-motion methods focus on generating motion from a discrete number of classes describing the motion. Some of these classes may include 'running,' 'jumping,' 'walking,' etc. For example, Action2Motion, \citep{Guo2020}, proposed a temporal VAE with Lie algebra-based pose representation. This approach was tested on the NTU-RGB+D dataset with 120 action classes and HumanAct12 with 12 course-grained action classes.
\citep{Petrovich2021} also used a VAE and evaluated their approach on the same dataset as \citep{Guo2020}. Unlike previous work that examined the loss in joint rotation or the loss in reconstruction in joint positions, the authors assessed the loss in reconstructed vertices of the SMPL body model \citep{Loper2015} and also opted for a 6D rotation representation \citep{Zhou2018_2} for training. \citep{Yang2018} proposed a pose sequence generative adversarial network (PSGAN) on labelled action classes to drive a human video generation model. PSGAN employed an encoder-decoder architecture with residual blocks and LSTM modules for temporal modeling.

In video-to-motion, \citep{Sun2021} introduced a framework to forecast human motion from past video frames. The core idea was utilizing action-specific memory banks to capture motion dynamics. The system first predicts the action class from the observed video frames and then queries the memory bank associated with the predicted action class to get relevant motion dynamics. Another work by \citep{Pavllo2018} achieves a slightly different goal, the authors proposed a method to extract the human motion data described in a video sequence using a temporal convolution model. Their work has served as a tool for powering optical motion capture devices.

More closely related to this study is motion generation from prior motion. In addition to the work \citep{Martinez2017, Pavllo2019} discussed in the previous section, \citep{Hernandez2018} presented STIM-GAN, designed to preserve temporal coherence between generated motion and introduced frequency-based metrics for accessing motion quality. \citep{Harvey2018} proposed a Recurrent Transition Network (RTN) to generate a realistic motion transition between different states. The core of RTN uses LSTM with initialized hidden states. \citep{Zang2020} proposed MoPredNet, a few-shot human motion prediction model that leverages deformable spatio-temporal convolution network (DSTCN), sequence masks, and a parameter generation module to adapt to new motion dynamics and make accurate long-term predictions. \citep{Qu2024} introduced FMP-OC, a framework that uses LLMs for predicting motion with a few shots without additional training. This framework bridges the gap between LLMs and numeric motion data. Unlike the classes of models discussed above, \citep{Holden2016} presented a convolutional autoencoder that takes in high-level parameters such as the trajectory of a character over terrain and the movement of the end effectors (hands and feet) to generate full body motion.

\subsection{Representing Rotation}
The choice of rotation representation is important in training neural networks as it directly affects the stability and accuracy of the learning process. \citep{Pavllo2019} highlighted two critical properties for rotation representation; continuity and interpolation. These properties rule out Euler angles as a suitable rotation representation. Exponential maps, another alternative to represent rotation, suffer the same fate. However, quaternions possess 3D interpolation properties, with only two quaternions encoding the same rotation. The authors presented an approach to resolve this ambiguity to ensure that the quaternion representation was continuous.

Upon further exploration of the rotation representation, \citep{Zhou2018_2} proposed the continuous rotation representation in higher dimensions. They proposed a method to represent rotation in 5D and 6D for any given 3D rotation. Another work by \citep{Xiang2020} discusses the effectiveness of higher-dimensional rotation representations such as 5D and 6D embedding. They revealed that even higher-dimensional representation could struggle in certain scenarios, particularly when dealing with point clouds with nontrivial rotational symmetry. The authors use ensembles of representation to mitigate these issues.

\section{Methodology}
Given a set of input poses \(\mathbf{c} \in \mathbb{R}^{L \times F}\) and respective pose indices \(\mathbf{c}_i \in \mathbb{R}^{L}\), \(F = 3 + J \cdot D\), \(D\) represents the number of dimensions of the joint representation (e.g., a quaternion or 6D rotation representation \citep{Zhou2018_2}), and \(J\) is the number of joints, the objective is to generate output motion sequence \(\mathbf{x} \in \mathbb{R}^{B \times F}\). \(L\) is the context length or the number of input poses. \(B\) is the block size or the length of the generated motion sequence. \(L\) ranges from \(1\) to \(B/2\), \(F\) is the dimensionality of each pose representation, which includes the 3D position of the root joint relative to the first frame in the motion sequence, joint features (e.g., rotations), which are encoded in \(D\) dimensions for each of the \(J\) joints. \(\mathbf{x}\) is ensured to retain the poses in \(\mathbf{c}\) at specified index \(\mathbf{c}_i\) and to follow the dynamics and patterns observed in \(\mathbf{c}\).

\subsection{Denoising Diffusion Model}
The proposed method uses a diffusion model for generating human motion. This process is inspired by the natural phenomenon of diffusion, where particles spread from regions of higher concentration to lower concentration until equilibrium is achieved. The exception here is that the aim is to learn the reverse process, from a state of high entropy (i.e. standard normal noise) to a state of low entropy (i.e. clean data). It consists of two sub-processes, the forward process and the reverse process.
The primary component of the forward process is the noise scheduler. Starting with clean data \(\mathbf{x}_0\) at \(t = 0\), Gaussian noise \(\mathcal{N}(\mu, \sigma^2)\) is gradually added to \(\mathbf{x}_0\). At each time step \(t\), \(\mu\), and \(\sigma\) are determined by the noise schedule. The goal is to transform \(\mathbf{x}_0\) into \(\mathbf{x}_t\) where \(\mathbf{x}_t\) is approximately a standard normal distribution \(\mathcal{N}(0, I)\) with high entropy. The reverse process aims to reverse the forward process by predicting clean data \(\mathbf{x}_0\), or the standard normal noise added. This stage is performed by a denoiser. In our case, at each time step, the denoiser takes in noisy data \(\mathbf{x}_t\) and predicts clean data \(\hat{\mathbf{x}}_0^t\). This step is done iteratively for the number of given time steps to generate the final clean data.

\subsection{Noise Schedule}
The choice of the beta schedule determines how noise is introduced during the forward process. Our method uses a noise schedule with a sample linear beta schedule \citep{Ho2020} among others including the cosine \citep{nichol2021improved} and the sigmoid \citep{Jabri2022} beta schedule. The noise schedule, denoted by \(\beta_t\), is defined as:

\begin{equation}
    \beta_t = \beta_{\text{min}} + t \cdot \frac{\beta_{\text{max}} - \beta_{\text{min}}}{T}
\end{equation}

where \(\beta_{\text{min}}\) and \(\beta_{\text{max}}\) are the minimum and maximum noise levels, respectively, and \(T\) is the total number of time steps.
With \(\beta_t\) defined, the forward diffusion process for any given time step $t$ can be described as in equation \ref{eq:noisy_data}. \(\bar{\alpha}_t\) represents the cumulative product of noise coefficients up to time step \(t\), and \(\epsilon \sim \mathcal{N}(0, I)\) is Gaussian noise.

\begin{equation}
    x_t = \sqrt{\bar{\alpha}_t} \, x_0 + \sqrt{1 - \bar{\alpha}_t} \, \epsilon
    \label{eq:noisy_data}
\end{equation}

\begin{equation}
    \bar{\alpha}_t = \prod_{s = 1}^t (1 - \beta_s)
\end{equation}

\subsection{Denoiser Network}
The denoiser network is built on the transformer architecture. It consists of two encoders. For each sampling step, a masked input encoded with the current time step is generated using \(c\) and \(c_i\) and passes through a stack of transformer encoder layers. The output, together with \(x_t\), which is initially standard normal noise, is fed into another stack of transformer encoder layers to predict clean motion \(\hat{x}_0^t\). If the current time step is greater than zero, \(x_{t-1}\), noisy data for the next denoising step are obtained from equation \ref{eq:noisy_data_t}. \(\hat{x}_0^t\) is the predicted clean motion at time step \(t\), \(\tilde{\beta}_t\) is the posterior variance at time step \(t\) and \(\epsilon \sim \mathcal{N}(0, I)\) denotes standard normal noise. This process continues until $T$ iterations are completed and the final clean motion \(\hat{x}_0^0\) is obtained.

\begin{equation}
    x_{t-1} = \hat{x}_0^t + \sqrt{\tilde{\beta}_t} \cdot \epsilon
    \label{eq:noisy_data_t}
\end{equation}

\subsection{Forward Kinematics}
To reconstruct the joint positions from individual local joint rotations and the global root position, we define a function \(FK(x; A(i), B)\) to perform the forward kinematic process. \(x\) is a tensor of rotation matrices representing local rotation for each joint. \(A(i)\) is a function that returns the kinematic chain of the target joint with index \(i\), this chain is an ordered
set of joints connecting the root joint to the target joint. \(B\) is an ordered set of position vectors for each joint defining the rest pose. For a joint at index \(i\), the position is calculated as shown in equation \ref{eq:fk}. \(j_p\) denotes the index of the parent to \(j\) in the kinematic chain \(A(i)\). For the case where \(j\) is the index of the root joint, we set \(B_{j_p}\) to a zero vector.
\begin{equation}
    p_i = \left( \prod_{j \in A(i)}
    \begin{bmatrix}
        x_j & 0 \\
        B_j - B_{j_p} & 1
    \end{bmatrix}\right)
    \begin{bmatrix}
        0 \\
        1
    \end{bmatrix}
    \label{eq:fk}
\end{equation}

\subsection{Training Objective}
The training objective at any time step is to minimize the error between \(x_0\) and \(\hat{x}_0^t\). We define a composite loss function \(l\) with five components for this. 

\begin{itemize}
    \item Model Loss \(l_g\): Measures the distance between the predicted clean motion \(\hat{x}_0\) at time step \(t\) and the ground truth motion \(x_0\).
    \begin{equation}
        l_g = \sqrt{\frac{1}{N} \sum_{i=1}^{N} \left(\hat{x}_0^{t(i)} - x_0^{(i)}\right)^2}
    \end{equation}
    \item Reconstruction Loss \(l_r\): Measures the distance between the reconstructed joint position of the predicted clean motion and the ground truth.
    \begin{equation}
        l_r = \sqrt{\frac{1}{N} \sum_{i=1}^{N} \left(FK(\hat{x}_0^{t(i)}) - FK(x_0^{(i)}) \right)^2}
    \end{equation}
    \item Context Loss \(l_c\): Measures the distance between the reconstructed joint positions of context keypoints in the predicted clean motion \(\hat{x}_0^t\) and the ground truth motion \(x_0\) for the context indices \(c_i\).
    \begin{equation}
        l_c = \sqrt{\frac{1}{|c_i|} \sum_{i \in c_i} \left(FK(\hat{x}_0^{t(i)}) - FK(x_0^{(i)}) \right)^2}
    \end{equation}
    \item Rotation Velocity Loss \(l_{r\_vel}\): Measures the distance between the rotation velocities of the predicted clean motion \(\hat{x}_0^t\) and the ground truth motion \(x_0\).
    \begin{equation}
        l_{r\_vel} = \sqrt{\frac{1}{N - 1} \sum_{i=1}^{N-1} \left((\hat{x}_0^{t(i + 1)} - \hat{x}_0^{t(i)}) - (x_0^{(i + 1)} - x_0^{(i)}) \right)^2}
    \end{equation}
    \item Position Velocity Loss \(l_{p\_vel}\): Measures the distance between the position velocities of the predicted clean motion \(\hat{x}_0^t\) and the ground truth motion \(x_0\).
    \begin{equation}
        l_{p\_vel} = \sqrt{\frac{1}{N - 1} \sum_{i=1}^{N - 1} \left((FK(\hat{x}_0^{t(i + 1)}) - FK(\hat{x}_0^{t(i)})) - (FK(x_0^{(i + 1)}) - FK(x_0^{(i)})) \right)^2}
    \end{equation}
\end{itemize}

\(FK\) is a function that computes the joint positions from joint orientation and root position using the forward kinematic process. The total loss is a sum of these components:
\begin{equation}
    l = l_g + l_r + l_c + l_{r\_vel} + l_{p\_vel}
\end{equation}

\section{Experiments}
\subsection{Dataset and Preprocessing}
Our experiment used five AMASS data sets \citep{AMASS:ICCV:2019}. This included CMU mocap \citep{cmuWEB}, BMLrub \citep{BMLrub}, DanceDB \citep{DanceDB:Aristidou:2019}, SFU \citep{SFU}, and MPI Limits \citep{PosePrior_Akhter:CVPR:2015}. We trained our model on CMU mocap, BMLrub, and DanceDB. We then evaluated the zero-shot performance of our model on SFU and MPI Limits. The CMU mocap dataset consists of 1983 motions from 96 subjects totaling 543 minutes of motion data. The dataset covers various human activities, including walking, running, dancing, and other everyday activities. The BMLrub dataset consists of 3061 motion sequences performed by 111 subjects covering 522 minutes of motion data. DanceDB consists of a wide range of dance styles and performances. It includes over 150 motions, with 20 subjects and a total duration of 203 minutes. SFU and MPI Limits are on the smaller side, with a total duration of 15 and 20 minutes, respectively. SFU consists of 44 motions with 7 different subjects, while MPI Limits includes 35 motion sequences with 3 subjects.
We executed a pipeline with three stages to preprocess the datasets. First, each motion sequence is downsampled to a target frame rate of 15 fps. This is done by selecting frames at an interval determined by the source and target frame rate. The second stage of the pipeline involves generating chunks of motion sequences for training purposes. Here, we split motion sequences into chunks of 75 frames each, representing 5 seconds. The final stage involves increasing the output of the previous stage. We extended the data size by rotating each motion sequence randomly twice around the vertical axis. We utilized a data split of 80-10-10 for training, validating and testing respectively.

\subsection{Experimental Setup}
In our experiment, we used 6D rotation representation \citep{Zhou2018_2} and two encoder transformers with the following configurations. Each encoder consists of six layers, an input dimension of 512, a feed-forward dimension of 2048, eight attention heads, and drop-out layers of 0.1 drop rate. The model was trained with a batch size of 128 and 300 diffusion time steps on a single NVIDIA P100 GPU provided by Kaggle.
\subsection{Evaluation Metrics}
To evaluate the effectiveness of our model in generating high quality and diverse motion sequences, we employ three different metrics: Frechet Inception Distance (FID), Diversity, and Multimodality. These metrics are discussed in the following.
\begin{itemize}
    \item Frechet Inception Distance (FID): FID was originally introduced to measure the performance of GANs on image generation tasks. It has continued to be used in general on generation tasks. It measures the distance between the extracted feature distribution of real and generated data. It works by comparing the mean and covariance matrix of the extracted features from both real and generated data. A lower FID score indicates the feature distribution of generated data is closer to real data and can be considered to be realistic and of high quality.
    \item Diversity: Diversity measures the variation in the model's generated output across various input condition. It can be calculated by averaging the pairwise distance between generated samples in feature space. High diversity indicates that the model can generate a wide variety of distinct outputs.
    \item Multimodality: This measures the ability of the model to generate varied motion sequences from similar input or context. This is useful for a generative model where there could be multiple valid outputs for the same input condition, and we want the model to capture the full range of this space. Calculated by measuring the average pairwise distance between generated samples that share the same input condition.
\end{itemize}
To extract the required features for computing these metrics, we trained an autoencoder to extract a feature embedding of \(\mathbb{R}^{256}\) from a given motion sequence, mapping \(f: \mathbb{R}^{B \times F} \to \mathbb{R}^{256} \).

\section{Results and Analysis}
We present the results of our evaluation as follows:
\subsection{Visual Evaluation of Generated Motion}
The following figures \ref{fig:sample1} to \ref{fig:sample4} illustrate the generated motion sequence output of our model. These images represent various frames of the generated sequences which illustrates motion dynamics over time. The generated frames have been downsampled for clearer visualization.

\begin{figure}[htbp]
    \centering
    \includegraphics[width=0.5\textwidth]{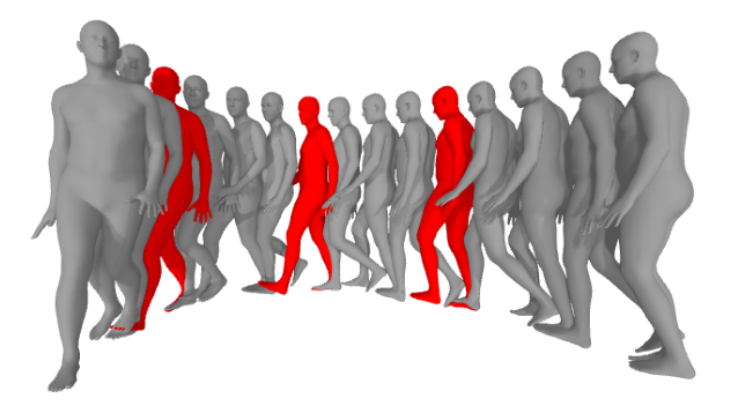}
    \caption{Sample output from our model on unseen input motion. The red body indicates the input poses. This has been downsampled for clearer visualization and the ratio of input poses to generated output is maintained.}
    \label{fig:sample1}
\end{figure}

\begin{figure}[htbp]
    \centering
    \includegraphics[width=0.5\textwidth]{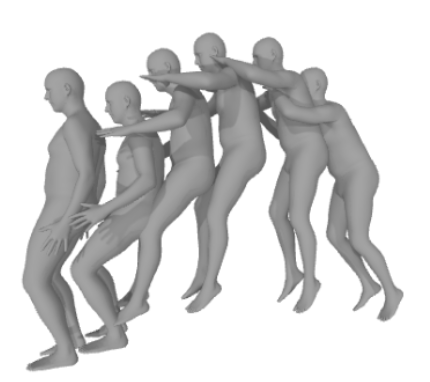}
    \caption{Sample output from our model on unseen input motion. This has been downsampled for clearer visualization. Each frame has been sufficiently spaced to prevent overlapping.}
    \label{fig:sample2}
\end{figure}

\begin{figure}[htbp]
    \centering
    \includegraphics[width=0.5\textwidth]{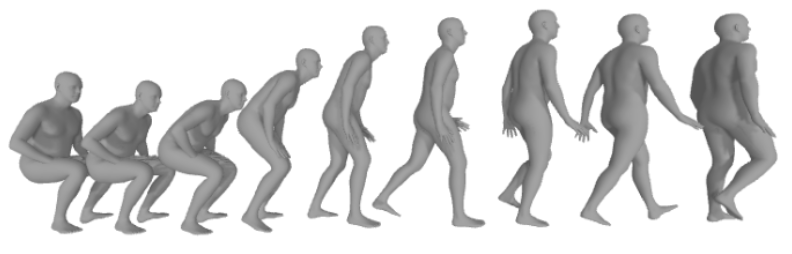}
    \caption{Sample output from our model on unseen input motion. Each frame has been sufficiently spaced to prevent overlapping.}
    \label{fig:sample3}
\end{figure}
\begin{figure}[htbp]
    \centering
    \includegraphics[width=0.9\textwidth]{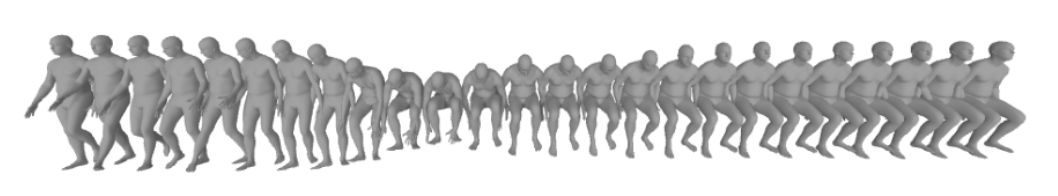}
    \caption{Sample output from our model on unseen input motion. Each frame has been sufficiently spaced to prevent overlapping. Root position and orientation are not visualized.}
    \label{fig:sample4}
\end{figure}

\subsection{Quantitative Evaluation Metrics}
In table \ref{tab:quantitative evaluation metrics}, we present the evaluation results of our model with input lengths of 10 and 20 across multiple datasets. The multimodality scores decrease as the number of input poses increases, due to stronger constraints imposed by the increased input length, reducing the ambiguity in the generated motion sequence. It is important to mention that this table does not include multimodality scores for real data as it is not grouped by identical motion frames.
\begin{table}[h!]
    \centering
    \caption{Evaluation metrics for different datasets}
    \begin{tabular}{llccc}
    \toprule
    \textbf{Dataset} & \textbf{Method} & \textbf{FID} \(\downarrow\) & \textbf{Diversity} \(\uparrow\) & \textbf{Multimodality} \(\uparrow\) \\
    \midrule

    \multirow{3}{*}{CMU}
    & Real    & \(0.026^{\pm 0.003}\) & \(2.788^{\pm 0.032}\) & - \\
    \cmidrule(lr){2-5}
    & Ours\(_{|c|=20}\) & \(0.392^{\pm 0.006}\) & \(2.991^{\pm 0.039}\) & \(0.165^{\pm 0.027}\) \\
    & Ours\(_{|c|=10}\)  & \(0.514^{\pm 0.009}\) & \(3.147^{\pm 0.033}\) & \(0.173^{\pm 0.019}\) \\
    \midrule

    \multirow{3}{*}{BMLrub}
    & Real    & \(0.015^{\pm 0.003}\) & \(2.545^{\pm 0.038}\) & - \\
    \cmidrule(lr){2-5}
    & Ours\(_{|c|=20}\) & \(0.253^{\pm 0.006}\) & \(2.625^{\pm 0.037}\) & \(0.097^{\pm 0.008}\) \\
    & Ours\(_{|c|=10}\)  & \(0.265^{\pm 0.007}\) & \(2.472^{\pm 0.028}\) & \(0.117^{\pm 0.011}\) \\
    \midrule

    \multirow{3}{*}{DanceDB}
    & Real    & \(0.073^{\pm 0.006}\) & \(3.610^{\pm 0.019}\) & - \\
    \cmidrule(lr){2-5}
    & Ours\(_{|c|=20}\) & \(0.716^{\pm 0.003}\) & \(3.751^{\pm 0.057}\) & \(0.362^{\pm 0.035}\) \\
    & Ours\(_{|c|=10}\)  & \(0.924^{\pm 0.002}\) & \(3.788^{\pm 0.015}\) & \(0.488^{\pm 0.048}\) \\
    \midrule

    \multirow{3}{*}{SFU}
    & Real    & \(0.199^{\pm 0.094}\) & \(3.033^{\pm 0.049}\) & - \\
    \cmidrule(lr){2-5}
    & Ours\(_{|c|=20}\) & \(0.742^{\pm 0.011}\) & \(3.139^{\pm 0.053}\) & \(0.209^{\pm 0.018}\) \\
    & Ours\(_{|c|=10}\)  & \(0.801^{\pm 0.005}\) & \(3.202^{\pm 0.038}\) & \(0.235^{\pm 0.039}\) \\
    \midrule

    \multirow{3}{*}{MPI Limits}
    & Real    & \(0.319^{\pm 0.087}\) & \(2.743^{\pm 0.046}\) & - \\
    \cmidrule(lr){2-5}
    & Ours\(_{|c|=20}\) & \(1.550^{\pm 0.001}\) & \(2.937^{\pm 0.040}\) & \(0.235^{\pm 0.031}\) \\
    & Ours\(_{|c|=10}\)  & \(2.268^{\pm 0.001}\) & \(3.0124^{\pm 0.068}\) & \(0.355^{\pm 0.042}\) \\
    \bottomrule
    
    \end{tabular}
    
    \label{tab:quantitative evaluation metrics}
\end{table}

\section{Conclusion}
In this study, we explored the use of diffusion models for motion stitching and in-betweening. We trained a transformer-based denoiser on datasets from AMASS and presented the output of sampling our model using the reverse diffusion process for 300 timesteps. Furthermore, we present the performance evaluation of our model using FID, Diversity, and Multimodality metrics. Some limitations of this study include the fixed output sequence length, motion generation from unrealistic input key poses and model performance degradation on smaller input context length. For future work, we aim to consider additional input conditions to capture more contextual information about the desired output motion sequence, for example, a textual description of the desired output motion. Conditioning on more contextual information is important to guide stitching and in-betweening on longer motion generation tasks.

\bibliographystyle{unsrtnat}

\bibliography{references}

\end{document}